# Unsupervised Outlier Detection in Audit Analytics:

# A Case Study Using USA Spending Data

**Buhe Li**

*Rutgers, The State University of New Jersey, Newark, USA*

*corresponding author: Buhe Li  buhe.li@rutgers.edu*

**Berkay Kaplan**

*Rutgers, The State University of New Jersey, Newark, USA*

berkay.kaplan@rutgers.edu

**Maksym Lazirko**

*Rutgers, The State University of New Jersey, Newark, USA*

mol12@business.rutgers.edu

**Aleksandr Kogan**

*Rutgers, The State University of New Jersey, Newark, USA*

kogan@business.rutgers.edu

**Acknowledgment:** The authors express their sincere gratitude to Dangyang Wei for her numerous critical contributions to this research.

Buhe Li, Rutgers, The State University of New Jersey, Newark, Rutgers Business School, Newark and New Brunswick, Department of Accounting and Information Systems, Newark, NJ, USA; Berkay Kaplan, Rutgers, The State University of New Jersey, Newark, Rutgers Business School, Newark and New Brunswick, Department of Information Technology, Newark, NJ, USA; Maksym Lazirko, Rutgers, The State University of New Jersey, Newark, Rutgers Business School, Newark and New Brunswick, Department of Accounting and Information Systems, Newark, NJ, USA; Aleksandr Kogan, Rutgers, The State University of New Jersey, Newark, Rutgers Business School, Newark and New Brunswick, Department of Accounting and Information Systems, Newark, NJ, USA

**Abstract**

This study investigates the effectiveness of unsupervised outlier detection methods in audit analytics, utilizing USA spending data from the U.S. Department of Health and Human Services (DHHS) as a case example. We employ and compare multiple outlier detection algorithms, including Histogram-based Outlier Score (HBOS), Robust Principal Component Analysis (PCA), Minimum Covariance Determinant (MCD), and K-Nearest Neighbors (KNN), to identify anomalies in federal spending patterns. The research addresses the growing need for efficient and accurate anomaly detection in large-scale governmental datasets, where traditional auditing methods may fall short. Our methodology involves data preparation, algorithm implementation, and performance evaluation using precision, recall, and F1 scores. Results indicate that a hybrid approach, combining multiple detection strategies, enhances the robustness and accuracy of outlier identification in complex financial data. This study contributes to the field of audit analytics by providing insights into the comparative effectiveness of various outlier detection models and demonstrating the potential of unsupervised learning techniques in improving audit quality and efficiency. The findings have implications for auditors, policymakers, and researchers seeking to leverage advanced analytics in governmental financial oversight and risk management.

**Keywords:**

*Outlier Detection; Machine Learning; Government Financial Reporting; Audit Analytics;*

# I. Introduction

Outlier detection in audit data analytics (ADA) ensures integrity and accuracy in financial reporting. As financial datasets grow exponentially in size and complexity due to advancements in information technology, traditional auditing methods are increasingly supplemented with sophisticated analytical techniques to identify anomalies that could indicate errors, fraud, or other irregularities (Fulcer et al., 2020).

Large-scale data environments highlight the challenges that emphasize the need for effective outlier detection methods in auditing. Traditional manual auditing techniques are time-consuming and susceptible to the risk of overlooking significant anomalies due to human error or data overload. This shortcoming has led to a paradigm shift toward integrating advanced computational techniques, such as machine learning and data mining, to enhance the detection capabilities of audit systems (Bao et al., 2020).

Unsupervised learning methods do not require predefined labels or outcomes and are well-suited for identifying outliers in audit datasets (Aggarwal, 2017). These methods analyze the inherent structures and patterns within the data to flag transactions that deviate significantly from established norms (Ahmed et al., 2016). These approaches offer advantages in scenarios where prior labeling is impractical or evolving dataset characteristics require adaptive methodologies.

The application of these methods, however, is challenging. One critical aspect is selecting a suitable model that balances sensitivity[1] and specificity[2]. The literature suggests a variety of algorithms, each with its strengths and limitations in handling different types of data irregularities (Akoglu et al., 2015). For instance, clustering-based techniques might excel in datasets with well-defined groupings, while distance-based methods might be preferable in more dispersed datasets.

---

1 Sensitivity - the ability to detect true outliers
2 Specificity - the ability to ignore non-outliers

Furthermore, integrating domain-specific knowledge into these models is vital for their success. Auditors' expertise and contextual understanding can guide the selection and tuning of parameters, improving the relevance and accuracy of the detection process (Bhuyan et al., 2013). This symbiosis between human judgment and algorithmic precision encapsulates the modern approach to auditing, where technology augments traditional practices.

This paper aims to expand on these foundations by proposing a framework that employs a hybrid approach—leveraging both unsupervised learning algorithms and expert insights. The framework intends to enhance the robustness of outlier detection in audit data analytics by incorporating adaptive learning mechanisms that can respond to dynamic changes in data patterns and audit environments.

In the following sections, we delve deeper into the specific methodologies employed, their implementation in the context of audit data analytics, and a discussion on the effectiveness of these methods based on recent case studies and audits.

**II. Background**

In the contemporary audit landscape, the increasing complexity and volume of data pose significant challenges for auditors, particularly in identifying transaction anomalies that may indicate risks such as fraud or errors. This paper seeks to address these challenges by harnessing unsupervised outlier detection methods within the U.S. Department of Health and Human Services (DHHS) audit context, a domain characterized by its vast data sets and critical need for accuracy and compliance.

Traditional methods, reliant on manual checks or rule-based systems, should be revised due to the sheer scale and subtlety of potential discrepancies. The unsupervised learning approach, particularly outlier detection, offers a promising avenue by automating the identification of anomalies without predefined labels. This method is efficient and capable of uncovering previously unnoticed irregular patterns, thereby enhancing the auditor's ability to focus on high-risk areas without the extensive labor and time typically required.

This research is motivated by the necessity to improve audit quality and efficiency in a data-driven era. By integrating and evaluating various outlier detection techniques, such as ensemble methods that combine multiple algorithms to improve detection accuracy, this study aims to develop a robust framework suitable for the complexities of the DHHS's financial environment. The anticipated outcome is a set of practical, scalable tools that can adapt to the dynamic nature of audit data, thus providing auditors with enhanced capabilities to detect and investigate outliers effectively. This enhancement not only aids in compliance and risk management but also contributes to the broader field of audit analytics by advancing methodologies that can adapt to other domains.

## III. Outlier Detection Methods Review

### 3.1 Overview of the popular outlier detection method

Outlier detection plays a crucial role across various domains, including finance, healthcare, and network security, where identifying anomalies can signal potential fraud, disease outbreaks, or cyber intrusions. This review encompasses four significant methodologies from the recent literature: Histogram-based Outlier Score (HBOS), Robust Principal Component Analysis (PCA), Minimum Covariance Determinant (MCD), and K-Nearest Neighbors (KNN). Each method offers unique advantages and contributes to the broader field of anomaly detection by addressing specific challenges associated with outlier identification in large and complex datasets. Table 1 summarizes what previous literature has done, and the following sub-sections go into further detail of these methods.

| Method | Description | Source |
|---|---|---|
| Histogram-based Outlier Score (HBOS) | Uses histograms to estimate the density distribution of features; identifies outliers based on deviations from these distributions | Goldstein et al. (2012) |
| Robust Principal Component Analysis (PCA) | Enhances traditional PCA to be less sensitive to outliers; improves data analysis quality in noisy datasets | Jackson and Chen (2004) |
| Minimum Covariance Determinant (MCD) | Seeks subset of dataset with minimal covariance determinant; provides resistant estimate against outliers in multivariate data | Hubert and Debruyne (2010) |
| K-Nearest Neighbors (KNN) | Uses distance-based approach to define outliers based on distance to nearest neighbors | Ramaswamy et al. (2000) |
| Local Outlier Factor (LOF) | Measures local deviation of data points relative to neighbors; assigns continuous outlier score | Breunig et al. (2000) |
| Cluster-Based Local Outlier | Combines clustering with outlier detection; calculates outlier factor based on cluster size and distance to | He et al. (2003) |
| Factor (CBLOF) | nearest large cluster | |
| Angle-Based Outlier Detection (ABOD) | Assesses variance in angles between different vectors of data points; addresses "curse of dimensionality" | Kriegel et al. (2008) |
| Isolation Forest (IF) | Uses tree structure to isolate anomalies; efficient for large, high-dimensional datasets | Liu et al. (2008) |
| Large Language Models (LLMs) | Processes unstructured textual data to identify semantic anomalies; complements traditional numerical methods | Not specified |

*Table 1: Outlier Detection Methods Overview*

### 3.1.1 Histogram-based Outlier Score (HBOS)

Goldstein et al. (2012) introduce HBOS, a fast and effective method for detecting outliers in large datasets. HBOS assumes independence between features and uses histograms to estimate the density distribution of each feature. Outliers are identified based on their deviations from these distributions.

**Main Findings:** HBOS is computationally efficient, making it suitable for real-time anomaly detection in large-scale applications.

**Contributions:** The method's simplicity and speed address the scalability challenges faced by traditional outlier detection techniques, especially in datasets with a large number of features.

### 3.1.2 Robust Principal Component Analysis (PCA)

Jackson and Chen (2004) discuss the application of robust PCA for identifying outliers in ecological data, where data integrity is often compromised by environmental noise and human error.

**Main Findings:** Robust PCA enhances the traditional PCA approach by being less sensitive to outliers, which improves the quality of the data analysis.

**Contributions:** This method contributes to the research stream by providing a more reliable tool for datasets heavily tainted with noise, thus broadening the applicability of PCA in outlier detection.

### 3.1.3 Minimum Covariance Determinant (MCD)

Hubert and Debruyne (2010) explore MCD, a robust estimator of multivariate location and scatter, for outlier detection. MCD seeks the subset of the dataset whose covariance determinant is minimal, providing a resistant estimate against outliers.

**Main Findings:** MCD is particularly effective in identifying outliers in multivariate data.

**Contributions:** The approach is pivotal for applications requiring robust statistical estimates as foundations for further analysis, significantly impacting fields that rely on accurate covariance estimation.

### 3.1.4 K-Nearest Neighbors (KNN)

Ramaswamy et al. (2000) propose an efficient algorithm for mining outliers using a distance-based approach, where outliers are defined based on the distance from a point to its nearest neighbors.

Main Findings: The method introduces a novel partition-based algorithm that significantly reduces computational requirements by focusing only on potential outlier regions.

Contributions: This technique is valuable in domains where the identification of local anomalies is critical, such as in network security and fraud detection.

### 3.1.5 Local Outlier Factor (LOF)

Breunig et al. (2000) introduced the Local Outlier Factor (LOF), which measures the local deviation of a given data point concerning its neighbors. This method is significant because it shifts from a binary notion of outliers to a continuum, assigning each data point a score representing the degree of the outlier.

**Main Findings:** LOF can identify regions in the dataset where points are considered outliers relative to their local neighborhoods.

**Contributions:** This technique brought a new perspective by focusing on the local aspects of data, which is especially useful in datasets where clusters vary in density.

### 3.1.5 Cluster-Based Local Outlier Factor (CBLOF)

He et al. (2003) proposed the Cluster-Based Local Outlier Factor (CBLOF), which classifies clusters into small clusters and large clusters and calculates the outlier factor based on the size of the cluster at a point and its distance to the nearest large cluster.

**Main Findings:** CBLOF introduces an approach that combines clustering with outlier detection, providing a measure that reflects the physical significance of an outlier.

**Contributions:** The integration of clustering dramatically reduces the computational complexity associated with traditional outlier detection methods, particularly in large datasets.

### 3.1.6 Angle-Based Outlier Detection (ABOD)

Kriegel et al. (2008) developed Angle-Based Outlier Detection (ABOD), which assesses the variance in the angles between different vectors connecting given points with all other points in the dataset.

**Main Findings:** ABOD addresses the "curse of dimensionality" by focusing on angular differences rather than distance metrics, which can degrade in high-dimensional spaces.

**Contributions:** This method offers a novel approach by considering angular measurements, thus providing a robust alternative to distance-based methods in high-dimensional data scenarios.

### 3.1.7 Isolation Forest (IF)

Liu et al. (2008) introduced Isolation Forest (IF), which isolates anomalies instead of profiling regular instances.

**Main Findings:** IF uses a tree structure to isolate anomalies, significantly reducing the need for a large sample size and enhancing computational efficiency.

**Contributions:** Its ability to handle high-dimensional datasets and its lower computational complexity makes it particularly effective for large datasets where traditional methods falter.

### 3.1.8 Large Language Models (LLMs) in Outlier Detection

Large Language Models (LLMs) represent a significant advancement in artificial intelligence and natural language processing, with potential applications in outlier detection for audit analytics. While not traditionally used for numerical outlier detection, LLMs offer unique capabilities that can complement and enhance existing methods.

**Main Findings:** LLMs can process and analyze unstructured textual data associated with financial transactions, identifying semantic anomalies that numerical methods might miss.

**Contributions:** The integration of LLMs in outlier detection introduces a new dimension of analysis, combining natural language understanding with traditional statistical approaches.

### 3.1.9 Comparison & Integration

These methods collectively enhance the outlier detection toolkit available to researchers and practitioners. While HBOS and KNN provide fast and effective solutions for large datasets, robust PCA and MCD offer resilience against data corruption. The LOF introduced the concept of local analysis, CBLOF integrated clustering for computational efficiency, ABOD offered a solution to the curse of dimensionality by using angle-based techniques, and IF revolutionized the field by using isolation rather than profiling, which is especially effective in large or complex datasets. An integrated approach combining these methodologies could leverage the strengths of each, providing a comprehensive outlier detection framework capable of addressing the diverse challenges presented by different data environments.

### IV. Research Questions

This study aims to investigate the effectiveness and integration of various outlier detection models in complex datasets, testing the findings from the aforementioned literature. The research objectives are twofold:

**Q1: What is the comparative effectiveness of various outlier detection models in identifying anomalies in complex datasets?**

This question invites an examination of how different outlier detection models perform across various complex datasets, which may include high-dimensional, noisy, or heterogeneous data types. Each model's effectiveness can be measured by its ability to accurately detect anomalies while minimizing false positives and false negatives. For instance, the Local Outlier Factor (LOF) excels in datasets with clusters of varying density but may struggle in uniformly distributed data. In contrast, Isolation Forest (IF) is noted for its efficiency in large datasets, and its performance does not degrade with an increase in dimensionality, making it suitable for high-dimensional spaces. Angle-Based Outlier Detection (ABOD) leverages angle variance to mitigate the curse of dimensionality, offering robustness in spaces where traditional distance measures fail. The comparative analysis could involve benchmarking these models against standard datasets that are publicly available, using metrics such as precision, recall, F1 score, and ROC-AUC to quantify and compare their performance.

**Q2: How does the integration of multiple detection strategies enhance the robustness and accuracy of outlier identification?**

Integrating multiple outlier detection strategies can potentially address the limitations of individual models by leveraging their respective strengths, leading to enhanced robustness and accuracy in outlier detection. For example, combining the density-based approach of LOF with the isolation mechanism of IF could allow for effective anomaly detection in mixed datasets featuring both dense clusters and sparse outliers. Similarly, integrating ABOD's angle-based approach with CBLOF's cluster-based method could improve performance in high-dimensional datasets by reducing the impact of irrelevant features and focusing on meaningful angular discrepancies. This integrative approach could be particularly beneficial in complex real-world scenarios where anomalies are not uniformly distributed or where they vary significantly in their nature. Experimental validation could involve applying these integrated models to diverse datasets and comparing their effectiveness against standalone models. Metrics such as detection time, scalability, and adaptability to new types of data anomalies could also be considered to evaluate the overall enhancement in robustness and accuracy.

**V. Dataset**

To test our research questions, this study utilizes the USA Spending dataset from the U.S. Department of Health and Human Services (DHHS), a comprehensive financial transaction database that provides detailed records of federal spending in the health sector. The publically available [3] dataset encompasses a wide range of financial activities, including contracts, grants, and other expenditures, offering a granular view of DHHS's fiscal operations.

The dataset covers all financial transactions conducted by the DHHS, providing a comprehensive overview of federal health spending. Also, each record includes detailed information such as transaction amounts, recipient details, and geographical distribution of funds. The dataset spans multiple fiscal years, allowing for longitudinal analysis of spending patterns and trends. Additionally, key variables, including transaction type, award amount, recipient information, program category, and associated dates, are also included. Overall, the dataset contains millions of records, reflecting the scale and complexity of DHHS operations.

This rich dataset presents a unique opportunity to apply advanced outlier detection techniques for identifying anomalous spending behaviors. By leveraging the dataset's comprehensive nature and granular detail, this study aims to detect potential irregularities that may indicate errors, fraud, or inefficiencies in federal health spending. The findings from this analysis have the potential to enhance financial transparency, improve resource allocation, and ultimately contribute to more effective public health outcomes through optimized financial governance.

---

3 https://www.usaspending.gov/agency/department-of-health-and-human-services?fy=2024

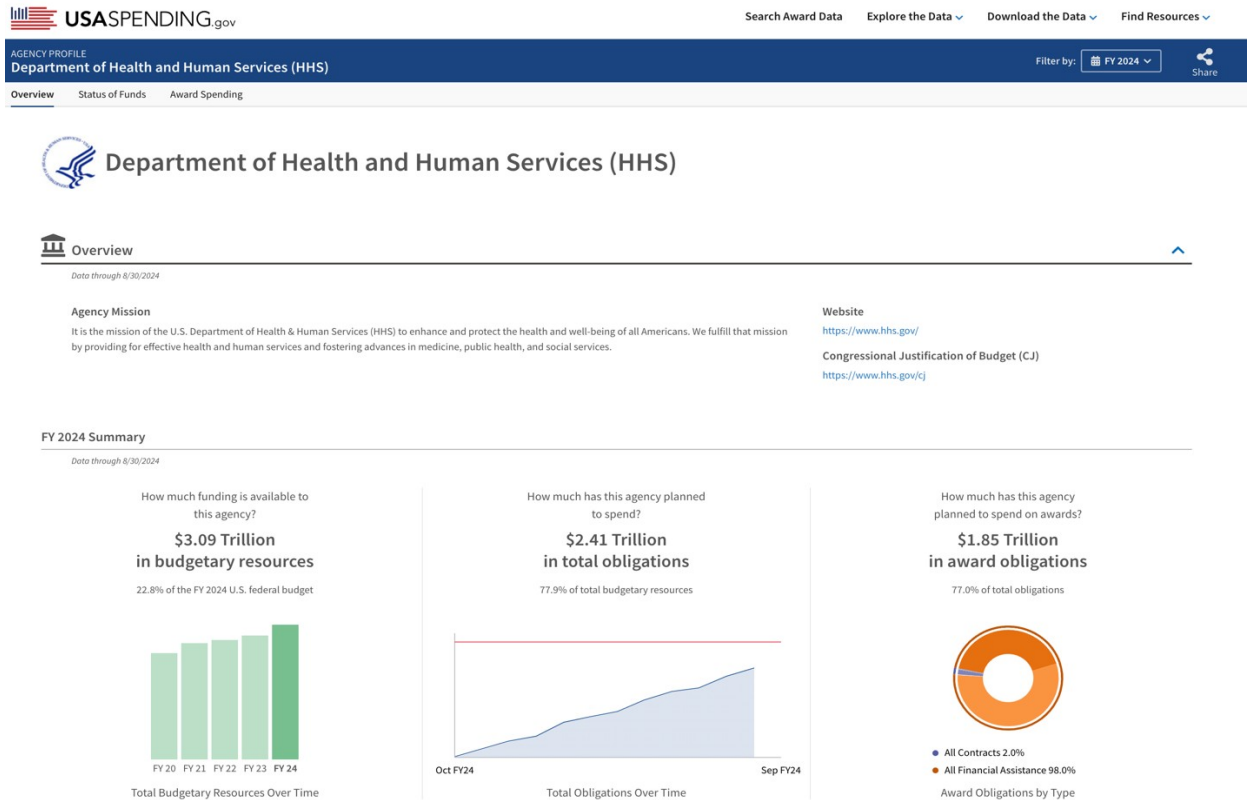

*Figure 1. DHHS Home Page Overview*

**VI. Framework**

We developed an outlier detection framework to identify high-risk transaction records for auditor examination during government financing. This comprehensive framework encompasses data preprocessing, the application of outlier detection algorithms, and result analysis. Ultimately, this structured approach facilitates robust conclusions, enhancing the efficiency and effectiveness of audit procedures.

## 6.1 Data Preparation

### 6.1.1 Pick financial data and calculate new columns

The dataset contains several columns related to financial transactions in government awards, possibly reflecting contracts or grants.

From the dataset 'FY(All)_075_Contracts_Delta_20240517.csv', we picked seven original columns and three calculated columns to form a new dataset.

These calculated fields are essential for financial analysis and forecasting, enabling stakeholders to understand the financial status and potential future liabilities or revenues associated with the awards. By comparing committed, obligated, and potential values, one can assess the financial health and operational efficiency of these government transactions. This comprehensive view aids in decision-making regarding continuing, expanding, or terminating contracts or grants based on their financial performance and projected value.

Table 2 provides a tabular view of the columns and the calculated fields:

| Column Number | Column Name | Description | Purpose |
|---|---|---|---|
| A | Federal Action Obligation | Initial amount legally committed to a specific action or contract | Represents the starting point of financial commitment |
| B | Total Dollars Obligated | Total amount obligated, including adjustments through | Reflects the current total financial |

|   |                                        | modifications and amendments                                            | commitment                                                        |
|---|----------------------------------------|-------------------------------------------------------------------------|-------------------------------------------------------------------|
| C | Total Outlayed Amount for Overall Award | Actual amount spent or disbursed under the award                       | Shows real cash flow as opposed to commitments                    |
| D | Base and Exercised Options Value       | Total value of base award plus exercised options                        | Indicates current contract value without modifications            |
| E | Current Total Value of Award           | Current total value after considering exercised options and modifications | Provides an up-to-date picture of the award's value            |
| F | Base and All Options Value             | Potential total value, including all options                            | Shows maximum possible value if all options are exercised         |
| G | Potential Total Value of Award         | Encompasses base, all options, and anticipated modifications            | Estimates the ultimate potential value of the award               |
| H | Net Obligation Difference(A-D) | Calculated as (Base and Exercised Options Value - Potential Total Value of Award) | Identifies discrepancies between commitments and potential value |
| I | Value Above Obligation(E-A) | Calculated as (Net Obligation Difference - Base and Exercised Options Value) | Monitors financial performance against commitments |
| J | Future Value Potential(G-F) | Estimates future financial exposure from remaining options and modifications | Projects potential additional value or exposure |

*Table 2: Data Tabular View*

### 6.1.2 Missing value treatment

Missing values in datasets like those collected by the US Treasury Department or the DHHS can occur for several reasons. Often, data may not be collected comprehensively due to oversight, errors in data entry, or gaps in the data collection process itself. Additionally, respondents may choose not to provide certain information, leading to systematic absences within the dataset. In some cases, technical issues such as malfunctioning data collection tools or loss of data during processing can also result in missing values. Addressing these gaps is crucial for maintaining the accuracy and reliability of analyses conducted with this data.

Using the mean to replace missing values in a dataset is a common technique, particularly advantageous for maintaining the overall statistical properties of the dataset. By filling NaN (Not a Number) values with the mean of their respective columns, the code helps to preserve the central tendency of the distribution for each numeric feature. This approach is beneficial when the data is normally distributed or when deviations caused by using the mean are minimal compared to other methods, such as using the median or mode.

One significant advantage of this method is its simplicity and the minimal impact it has on the variance and mean of the dataset. It avoids the introduction of bias that might occur with more arbitrary imputation methods, like filling missing entries with a constant value. Additionally, using the mean is computationally straightforward. It doesn't require complex calculations or understanding of the data's distribution, making it a practical choice for initial analyses where the priority is to quickly address missing values without introducing substantial biases.

However, it's important to note that this method assumes that the missing data is missing completely at random (MCAR). If this assumption does not hold, using the mean can lead to biased estimates, mainly if the missingness is related to the value itself (i.e., not random). In such cases, more sophisticated methods of imputation might be required to model the underlying data structure accurately.

### 6.1.3 Data Normalization

Data normalization is helpful for feature scaling; scaling itself is necessary for machine learning algorithms. This is because specific algorithms are sensitive to scaling. Let's look at it in more detail.

Distance algorithms like KNN, K-means, and SVM use distances between data points to determine their similarity. They're most affected by a range of features. Machine learning algorithms like linear regression and logistic regression use gradient descent for optimization techniques that require data to be scaled. Having similar scale features can help the gradient descent converge more quickly towards the minima. On the other hand, tree-based algorithms are not sensitive to the scale of the features. This is because a decision tree only splits a node based on a single feature, and this split is not influenced by other features.

Min-max scaling is a popular data normalization technique that adjusts the scale of your data to a predefined range, typically between 0 and 1. This method ensures that all features contribute equally to the analysis by eliminating the bias caused by the different units or scales in raw data. The formula used for min-max scaling is:

$$x' = \frac{x - min(x)}{max(x) - min(x)}$$

Here, x represents an original value, min is the smallest value in the dataset for that feature, and max is the largest value. The result, x′, is the normalized value that falls within the new specified range.

By rescaling the values, min-max normalization ensures that each feature contributes proportionately to the final distance calculations in algorithms that are sensitive to variations in magnitude, such as k-means clustering or k-nearest neighbors. This method is advantageous when you are working with parameters on different scales and you need a bounded range; however, it can be sensitive to outliers since the presence of extreme values can skew the range significantly.

## 6.2 Machine Learning Library

PyOD (Python Outlier Detection) is an open-source[4] Python library that provides a comprehensive and scalable toolkit for detecting outlying objects in multivariate data. It is explicitly designed for anomaly detection and offers a wide range of algorithms, from classical approaches like k-nearest neighbors (KNN) and outlier ensembles to state-of-the-art methods such as isolation forests and clustering-based outlier detection.

The library is well-suited for both academic research and commercial applications, featuring a unified API that simplifies the process of experimenting with different models to find the most effective solution for a particular dataset. PyOD also includes utility functions for evaluating and comparing the performance of various algorithms, using standard metrics like precision, recall, and ROC-AUC scores.

Additionally, PyOD is designed with efficiency in mind, supporting both multi-threading and serialization of models for easy deployment. This makes it an attractive option for handling large-scale datasets in real-world settings. Whether you are a data scientist, researcher, or machine learning enthusiast, PyOD provides a robust toolkit to help you efficiently detect and handle anomalies in complex datasets.

## 6.3 Application of Outlier Detection Algorithms

### 6.3.1 Histogram-based Outlier Score (HBOS)

HBOS is an efficient univariate method that calculates the anomaly degree by assuming independence between features. It is particularly effective for large, high-dimensional datasets because of its computational efficiency. By constructing histograms and calculating outlier scores based on the frequency distribution of the variables, HBOS is adept at identifying outliers in distributions that deviate from normality.

---

4 https://pyod.readthedocs.io/en/latest/pyod.models.html#

### 6.3.2 Principal Component Analysis (PCA)

PCA is primarily used for dimensionality reduction, capturing the major variance within the data through a lower-dimensional subspace. Outliers are identified by their reconstruction errors from this subspace. Although PCA is valuable for revealing underlying patterns that indicate abnormal behavior, its effectiveness can diminish with an increase in noise levels within the data.

### 6.3.3 Minimum Covariance Determinant (MCD)

MCD is a robust statistical technique that estimates the covariance matrix and means of the dataset by minimizing the determinant of the covariance matrix derived from a subset of the data presumed to be outlier-free. This makes MCD particularly resistant to outliers, thus effective in detecting multivariate anomalies.

### 6.3.4 k-Nearest Neighbors (KNN)

KNN determines outliers based on the proximity of neighboring points, identifying anomalies by the greater average distance to their k-nearest neighbors. This method is non-parametric and flexible, making it well-suited for datasets with non-linear variable relationships.

### 6.3.5 Local Outlier Factor (LOF)

LOF detects outliers by measuring local density deviations compared to neighbors. This method is proficient in spotting anomalies that may not stand out globally but are conspicuous in a local context. By focusing on local density around a specific data point relative to its immediate neighbors, LOF can pinpoint outliers in regions of varying density.

### 6.3.6 Cluster-Based Local Outlier Factor (CBLOF)

In CBLOF, data is classified into small and large clusters through a clustering algorithm. Outliers are identified by comparing points in smaller clusters against those in larger clusters, effectively distinguishing groups of outliers from regular observations, particularly in datasets with clear groupings.

### 6.3.7 Autoencoders

Autoencoders, leveraging neural network architectures, are designed to learn an identity function that closely reconstructs non-outliers while generating larger reconstruction errors for outliers. This approach is powerful for handling datasets with complex inter-feature relationships, as it learns nonlinear transformations.

### 6.3.8 Isolation Forest

Isolation Forest identifies anomalies by isolating them rather than modeling normal data points. It randomly selects a feature and a split value between the maximum and minimum values of that feature, recursively partitioning the data. Since anomalies are few and distinct, they tend to be isolated earlier in the process, making this method highly efficient and scalable, suitable for large datasets.

## 6.4 Outlier Detection Algorithms Parameter Utilized

In machine learning, the way algorithms perform can be greatly affected by how we choose and adjust their settings, known as hyperparameters. Each hyperparameter controls a different aspect of the algorithm's behavior, which can significantly impact the results. Properly tuning these settings through model validation and techniques like grid search can greatly improve the accuracy of the model, help avoid overfitting, and ensure the model performs well on new, unseen data.

In our initial tests, we used the default settings from the PyOD library for each algorithm. After this first run, we identified the top five suspicious records from each algorithm, labeling them as outliers. The frequency of their appearances is depicted in the table below:

| Award Number | Frequency of Award Number in Top 5 Lists |
|:---:|:---:|
| 33870 | 6 |
| 33868 | 4 |
| 33869 | 4 |
| 66 | 4 |
| 9797 | 4 |
| 67 | 2 |
| 14402 | 2 |
| 28222 | 2 |
| 714 | 2 |
| 68 | 2 |
| 33848 | 1 |
| 33854 | 1 |
| 11626 | 1 |
| 25668 | 1 |
| 28345 | 1 |
| 22877 | 1 |
| 33856 | 1 |
| 22527 | 1 |

*Table 3: Top 5 Award Number Frequency*

In the future phase of this project, we might receive officially labeled outlier records from the U.S. Department of Treasury. For now, we are using the suspicious records we've identified as labeled outliers to train our model and fine-tune the parameters through experimental design.

Initially, we introduced these labeled outlier data into the HBOS algorithm, adjusting the HBOS parameter, n_histograms, to determine which value best identifies the labeled outliers. The parameter n_histograms specifies the number of bins used in constructing histograms, which are critical for the algorithm's performance. Finer granularity in these histograms, achieved by increasing the number of bins, enhances the sensitivity of outlier detection. It allows for a more detailed representation of data distributions, improving the detection of subtle deviations in data density. However, setting too many bins can cause overfitting, particularly in cases of sparse data, as the model becomes overly sensitive to minor data variations. Properly calibrating n_histograms is crucial because it directly affects the algorithm's thresholding ability and overall effectiveness. Our experiments showed that setting n_histograms to 20 yielded the best results, with HBOS correctly predicting 100% of the labeled outliers.

For other algorithms, we obtained the following results:

- PCA with n_components set to 2 achieved a 94% accuracy rate.

- MCD with support_fraction set to 0.01 detected 100% of outliers.

- KNN with n_neighbors set to 5 identified 100% of outliers.

- LOF using n_neighbors of 5 reached a 94% accuracy rate.

- CBLOF with n_clusters set to 5 also achieved a 94% detection rate.

- An autoencoder (AE) with hidden neurons set as [64, 32, 32, 64] correctly identified 100% of outliers.

- Isolation Forest (IF) with n_estimators set to 100 showed a lower performance, detecting only 40% of the outliers.

These results highlight the importance of parameter tuning in enhancing the predictive accuracy of various outlier detection models. The experiment parameter can be found in the following table:

| Method | Hyperparamer 1 | Best Performed Parmeter | Prediction Correction Rate(%) |
|---|---|---|---|
| HBOS | n_histograms: [5, 10, 15, 20, 50, 100] | 20 | 100 |
| PCA | n_components: [0, 1, 2, 3, 4, 5, 6, 7] | 2 | 94 |
| MCD | support_fraction: [0, 0.01, 0.1, 0.2, 0.3, 0.4, 0.5, 0.6, | 0.01 | 100 |

| KNN | n_neighbors: [1, 3, 5, 10, 20, 30, 40, 50] | 5 | 100 |
|---|---|---|---|
| LOF | n_neighbors: [1, 3, 5, 10, 20, 30, 40, 50] | 5 | 94 |
| CBLOF | n_clusters: [2, 5, 10] | 5 | 94 |
| AE | hidden_neurons: [[64, 32, 32, 64], [32, 16, 16, 32], | [64, 32, 32, 64] | 100 |
| IF | n_estimators: [50, 100, 200, 300, 400, 500] | 100 | 40 |

*Table 4: Hyperparameter of each Algorithm*

**6.5 Ordering & Ranking Outlier Score**

**6.5.1 Ordering Outlier Score**

Once we've normalized the outlier scores from each algorithm, the next step involves ordering and ranking these algorithms to better understand their performance.

We utilize ordering to visually represent the data. For this, we first arrange the outlier detection scores from each algorithm in descending order. Each normalized outlier score is then assigned a numerical value, with the highest score labeled as No.1 and the lowest score, typically zero, marked as the last. Below, we use the normalized HBOS scores as an example. It is important to note that identical HBOS scores will have unique ordinal numbers. For instance, multiple instances of an HBOS score of 0.46 are differentiated by sequential ordering, such as 13 and 14, as demonstrated in the table.

## 6.5.1 Ranking Outlier Score

Ranking, on the other hand, takes a slightly different approach. The rank for each score is calculated based on a data ensemble method, where the same outlier scores receive a ranking that is the average of their positions. For example, if the HBOS score of 0.46 occurs at positions 13 and 14, both occurrences would be assigned a rank of 13.5. This method ensures that the ranking reflects the relative position of each score while accounting for ties by averaging their positions. Below is how these concepts are applied in the table format.

| Award Number | Norm_HBOS | Norm_HBOS_Ordering | Norm_HBOS_Ranking |
|---|---|---|---|
| 68 | 1.00 | 1 | 1 |
| 67 | 0.90 | 2 | 2 |
| 69 | 0.77 | 3 | 3 |
| 66 | 0.68 | 4 | 4 |
| 72 | 0.66 | 5 | 5 |
| 70 | 0.65 | 6 | 6 |
| 71 | 0.63 | 7 | 7 |
| 81 | 0.53 | 8 | 8 |
| 92 | 0.52 | 9 | 9 |
| 76 | 0.51 | 10 | 10 |
| 2174 | 0.47 | 11 | 11 |
| 2081 | 0.47 | 12 | 12 |
| **248** | **0.46** | **13** | **13.5** |
| **179** | **0.46** | **14** | **13.5** |
| 798 | 0.45 | 15 | 15 |
| 172 | 0.45 | 16 | 16 |
| 507 | 0.44 | 18 | 17.5 |
| 131 | 0.44 | 17 | 17.5 |
| 74 | 0.43 | 19 | 19 |
| 2156 | 0.42 | 30 | 25 |
| 357 | 0.42 | 29 | 25 |
| 239 | 0.42 | 28 | 25 |
| 225 | 0.42 | 27 | 25 |
| 216 | 0.42 | 26 | 25 |
| 180 | 0.42 | 25 | 25 |

*Table 5: HBOS Ordering and Ranking Example*

This table highlights the distinction between ordering, which assigns a unique sequential number to each score, and ranking, which averages the positions of identical scores for a fair and balanced evaluation. This approach not only provides clarity but also facilitates the comparison of performance across different algorithms based on a unified scaling system. We'll also apply this ordering and ranking method to the rest of the outlier score.

## 6.6 Process of Parameter Adjustment and Award Number Calculation

After adjusting the parameters based on the performance with labeled outliers, each algorithm is run to identify the top 5 outliers, often referred to as "award numbers." These are essentially the records that are most frequently flagged as outliers across multiple tests or parameter settings.

The subsequent step involves collating these award numbers from each algorithm and calculating their frequency of appearance. This helps in identifying the most consistently flagged outliers across different algorithmic implementations and parameter configurations.

Below is a table that illustrates the frequency of appearance for these top 5 outlier award numbers. This format effectively displays which outliers are most commonly detected across various algorithm settings, offering valuable insights into the nature of the data and the robustness of the algorithms used.

| Award Number | Frequency of Award |
|:---:|:---:|
| 66 | 5 |
| 33870 | 5 |
| 68 | 4 |
| 33869 | 4 |
| 9797 | 4 |
| 67 | 2 |
| 33856 | 2 |
| 33868 | 2 |
| 14402 | 2 |
| 28222 | 2 |
| 714 | 2 |
| 69 | 1 |
| 33513 | 1 |
| 33861 | 1 |
| 28345 | 1 |
| 22877 | 1 |
| 22527 | 1 |

*Table 6: Adjusted Parameter Top 5 Award Number Appear Frequency*

Based on this table, we create a bar chart to visualize the distribution of the top 5 outlier award number frequency. The bar chart is a great way to quickly assess which outliers are flagged most frequently across different algorithm executions.

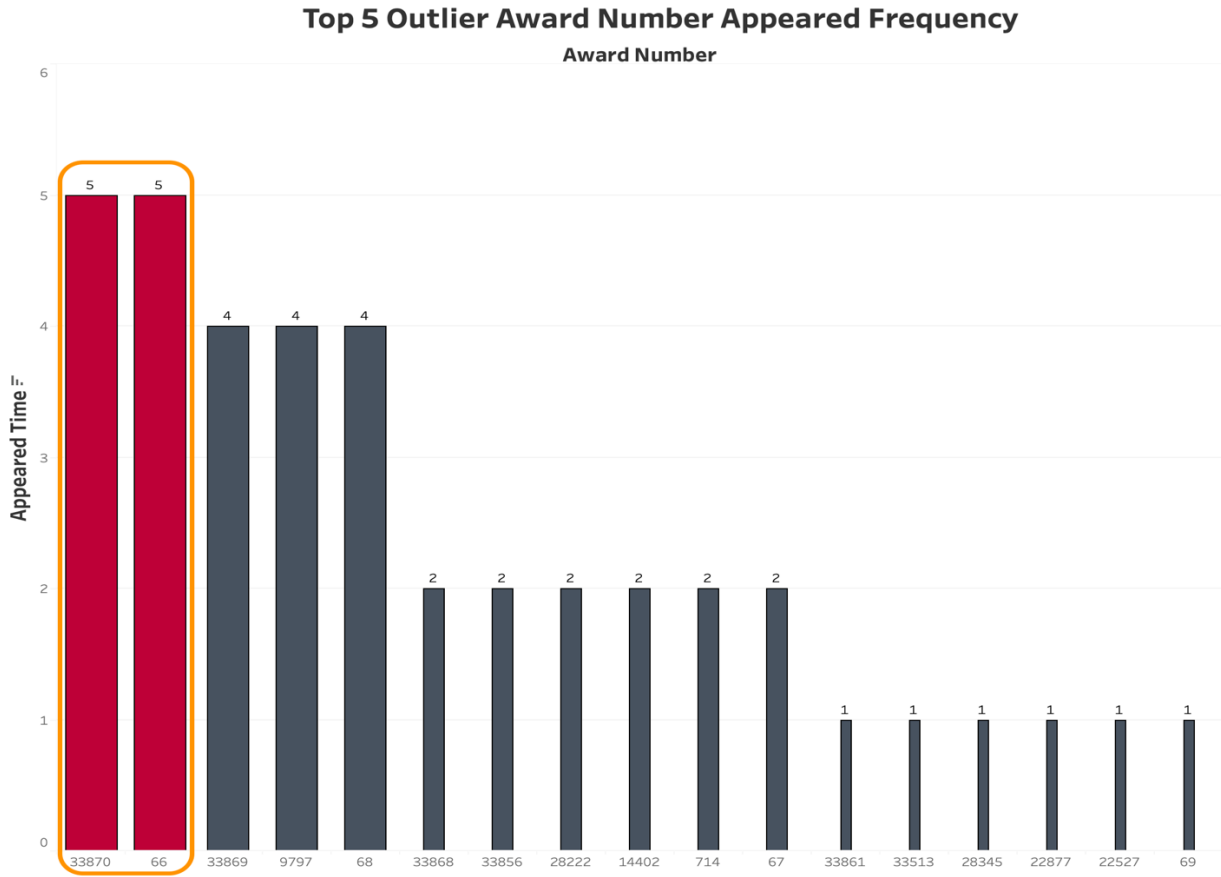

*Figure 2. Top 5 Outlier Award Number Appeared Frequency*

In this bar chart, the Y-axis (Vertical) represents the frequency of appearance for each award number, i.e., how many times each outlier has been flagged by the algorithms. X-axis (Horizontal) lists the award numbers themselves. Each bar in the chart corresponds to one of the top outlier award numbers identified.

The bar chart will display bars of varying heights, corresponding to how often each of the top 5 award numbers has appeared across tests. The height of each bar will directly reflect the number of times each outlier has been flagged, making it easy to spot the most frequently occurring outliers briefly. This visual representation makes it immediately apparent which outliers are most recurrent and may warrant further investigation due to their frequent identification as anomalies.

From the result of the bar chart, award numbers 66 and 33870, which appear 5 times, can be identified as highly subspinous outliers because they appeared most frequently under each algorithm.

### 6.7 Data Ensemble – Using Average Outliers Score and Average Ranking

The Data Ensemble method combines previous outlier scores and ranking values to compute both the average outlier score and the average ranking value. This aggregation helps in synthesizing insights from multiple runs or parameter settings of the algorithms, providing a holistic view of their performance.

From previous analyses, we identified award numbers 66 and 33870 as the most frequently flagged outliers across various algorithms. In this step, we will validate whether these award numbers indeed represent highly suspicious outliers by utilizing a data ensemble method.

### 6.7.1 Calculation and Visualization of Average Outlier Score

We first calculate the average outlier score for each award number by summing the scores from the HBOS algorithm to the IF algorithm and then dividing by eight the total number of algorithms used in this project.

Next, we normalize these average scores using the min-max method to ensure comparability. The normalized average outlier scores are then ordered in descending order.

We plot these normalized scores on a graph, connecting the scatterplot data points with a line. The y-axis represents the normalized average outlier score, and the x-axis represents the ordering of these scores. The top five award numbers are highlighted for emphasis.

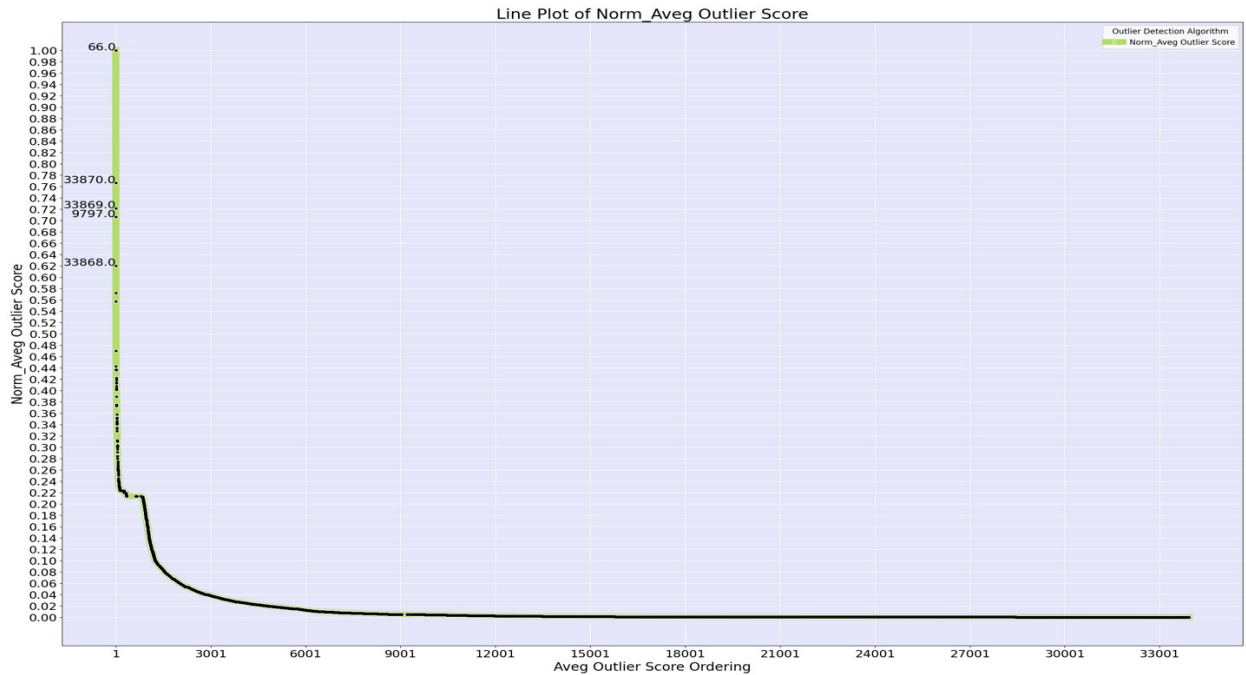

*Figure 3. Line Plot of Norm_Average Outlier Score*

**6.7.2 Calculation and Visualization of Average Ranking Score**

We then calculate the average ranking for each award number by summing up their rankings from each algorithm, dividing by eight, and normalizing these values using the min-max method.

To align the trend of the normalized scores with the ranking, we use the transformation '1-normalized average ranking score'. This inversion ensures that a higher normalized outlier score, which indicates a higher degree of outlierness, corresponds to a higher-ranking position.

On a new graph, the y-axis represents the '1-normalized average ranking score', and the x-axis shows the ordering of these transformed scores. Due to visual clarity and the challenge of presenting the top five together, only the award number with the highest '1-normalized average outlier score is displayed.

These steps help us confirm the true outlier status of the flagged award numbers by cross-verifying through both average outlier scores and rankings. The graphs serve as a visual aid to effectively communicate these trends, ensuring that the most suspicious outliers are distinctly highlighted. This rigorous validation process solidifies the reliability of our findings, enhancing the credibility of the analysis.

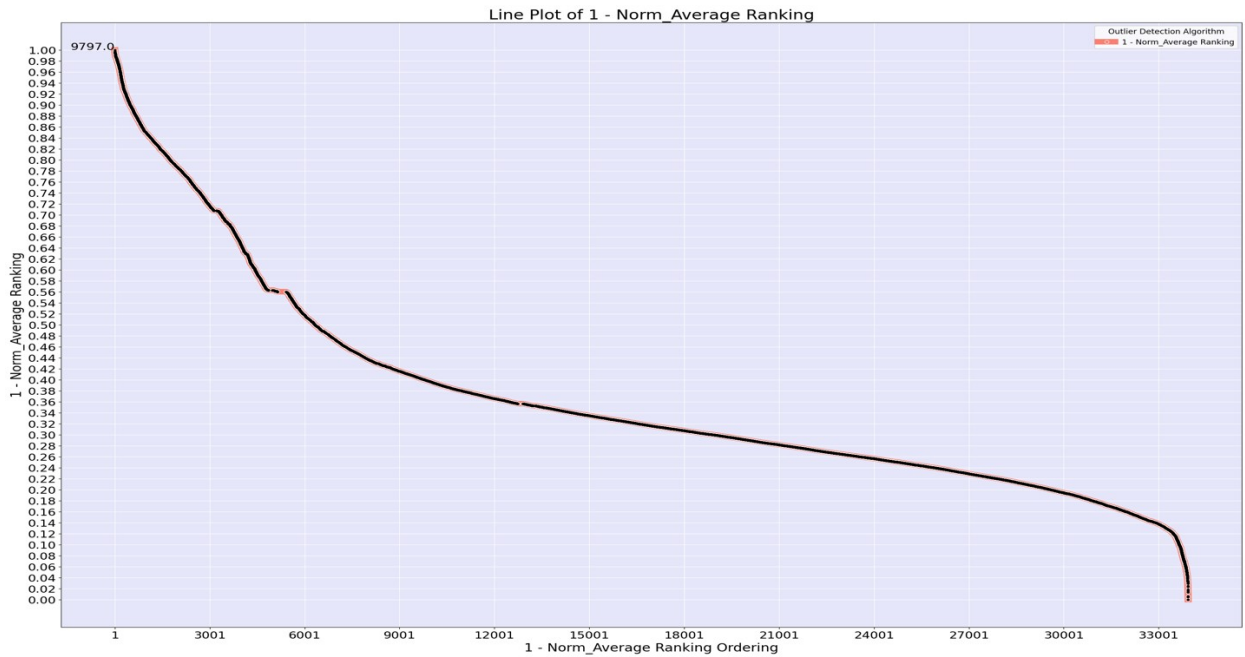

*Figure 4. Line Plot of 1 - Norm_Average Ranking Score*

| Award Number | Norm_Aveg Outlier Score | 1 - Norm_Average Ranking |
|:---:|:---:|:---:|
| 1 | 66 | 2014 |
| 2 | 67 | 2027 |
| 3 | 68 | 2469 |
| 4 | 2081 | 2952 |
| 5 | 714 | 1064 |

*Table 6: Top 5 Award Number Appeared Frequency*

This table effectively showcases the top five award numbers based on normalized outlier scores and the 1-normalized outlier ranking score. From our previous analysis in step 6.6, award numbers 66 and 33870 were identified as highly suspicious outliers. Notably, in this data ensemble phase, award number 66 is flagged once again, reinforcing its status as a significant anomaly.

Given its repeated identification as an outlier, we must conduct further investigation into the contracts associated with award number 66. This consistent flagging across different methodologies not only underscores the robustness of our analytical approach but also highlights the potential risks or irregularities associated with this award number. A deeper examination will help us understand the underlying factors contributing to this anomaly and guide appropriate actions to address any issues uncovered.

**6.8 Suspicious Data Explanation**

The detailed contract screenshot for award number 66 from the DHHS database has provided valuable insights into why this particular award was flagged as a highly suspicious outlier. The contract was issued to 1st Choice, LLC, a professional management consulting firm, to deliver administrative support, waste management, and remediation services. Upon further examination of the financial specifics provided in the "Award Amounts and Summary of All Federal Accounts Used by this Award" section, several anomalies were noted that likely contributed to the outlier status identified by our algorithms.

The contract details reveal that the Obligated Amount, Current Award Amount, Potential Award Amount, and Total Funding Obligated are all listed at an identical figure of $96,270. This uniformity is unusual and deviates from the typical financial progression expected in ongoing contracts.

Expected Financial Progression in Normal Contracts: Normally, for an ongoing project, the financial values should display a progression:

- Obligated Amount should be the initial committed funds and, thus, typically the lowest figure.

The current Award Amount represents the revised sum that reflects current contract modifications, which should be higher than the Obligated Amount if additional approvals or expansions have occurred.

- Potential Award Amount indicates the maximum possible funding considering all potential modifications and approvals, expected to be the highest value.

The identical figures across all key financial metrics in this contract raise red flags. The lack of differentiation between the Obligated, Current, and Potential Award Amounts suggests that the contract may not be progressing as expected, or there may be errors or misrepresentations in the reporting of financial data.

Given these findings, it is crucial to conduct a more thorough investigation into the activities and financial reporting of 1st Choice, LLC, related to this contract. This should include a detailed audit of the contract's progression, amendments, and the corresponding financial documentation to ensure transparency and compliance with federal contracting regulations. Such steps will help determine the nature of the anomalies and address any issues that might pose risks to the integrity of the contracting process and the effective use of public funds.

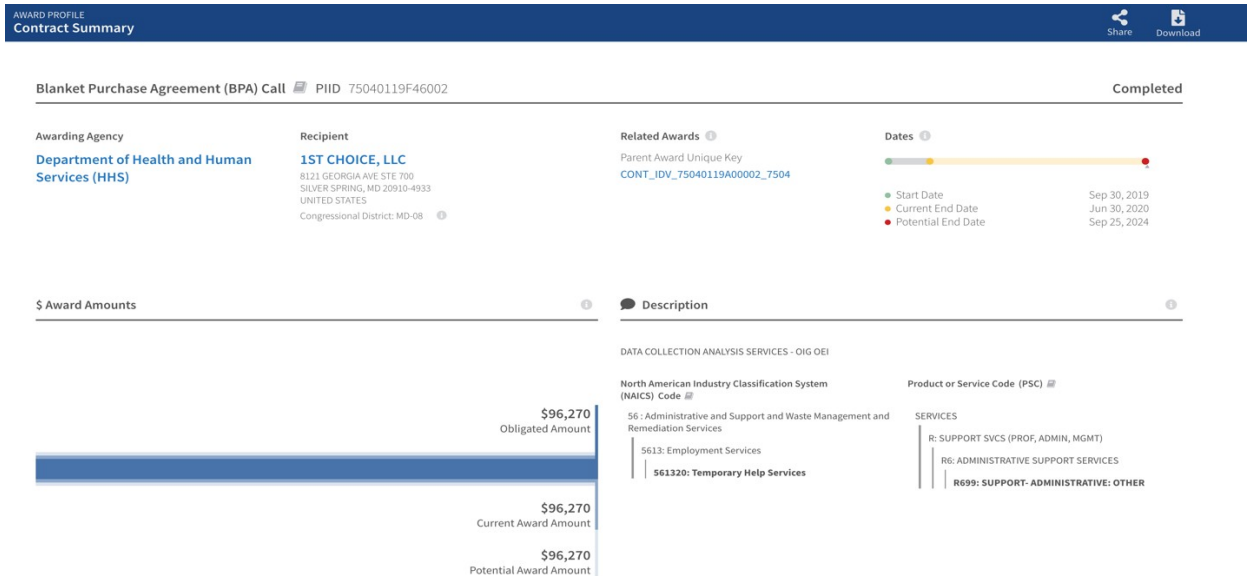

*Figure 5. Screenshot of Award Number 66 Overview 1*

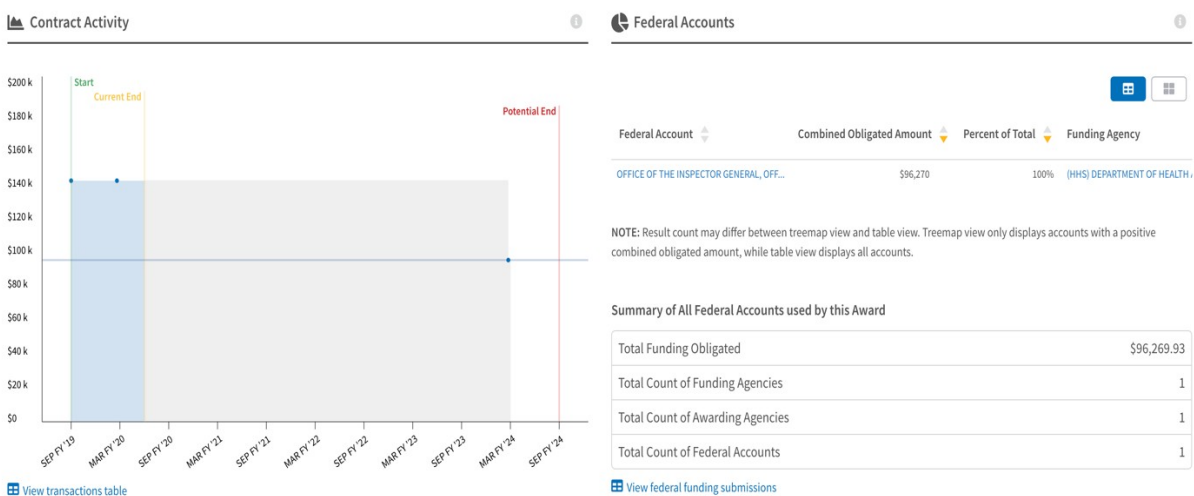

*Figure 6. Screenshot of Award Number 66 Overview 2*

## VII. Challenges

### 7.1 Missing Labeled Outlier Data

One of the primary challenges in this project is the absence of predefined outlier labels from the U.S. Treasury Department, which significantly complicates the validation of the outlier detection methods employed. Typically, outlier detection in supervised learning scenarios benefits from a labeled dataset where the anomalies are identified, allowing for straightforward training and evaluation of models. However, in unsupervised scenarios such as this, where labels are not provided, determining the effectiveness and accuracy of the detection methods becomes inherently complex.

Without these labels, it is challenging to assess the true positive rate of detected outliers—i.e., the proportion of actual outliers correctly identified by the model. Similarly, it becomes difficult to measure the false positive rate or the incidence of normal instances being incorrectly classified as outliers. This lack of labeled outliers forces reliance on indirect metrics of performance, such as consistency across multiple detection techniques or the stability of outlier scores under perturbations of the dataset.

Another significant challenge is the risk of model bias. In the absence of labels, there is a potential bias towards the majority pattern seen in the data, which may lead to overlooking subtle yet significant anomalous patterns. This is particularly problematic in financial datasets where anomalies can be indicative of critical, irregular transactions, such as fraud or errors in data entry. Additionally, the diverse nature of outlier detection algorithms—from distance-based methods like KNN and LOF to more complex models like Autoencoders and Isolation Forests—introduces variability in what each model perceives as an outlier. This heterogeneity can lead to a lack of consensus on which instances are truly anomalous, complicating the interpretation and integration of results from different methods.

Lastly, the challenge of parameter selection in unsupervised learning scenarios is non-trivial. Many outlier detection algorithms require the setting of parameters (e.g., the number of neighbors in KNN, the number of clusters in CBLOF, or the contamination factor in Isolation Forest), which significantly affects their performance. In the absence of outlier labels, optimizing these parameters often relies on heuristic approaches or assumptions about the data distribution, which may not always be valid or optimal for the specific dataset in question.

These challenges highlight the need for robust methodological approaches and careful consideration of model assumptions and parameters to ensure that the detection of outliers is both accurate and meaningful in the absence of direct guidance from labeled data.

## 7.2 Data Quality

During the practice of this project, we encountered significant challenges regarding data quality, particularly with the USAspending.gov dataset. We get some communications from the Governmental Accounting Standards Board(GASB), shared insights that echo our experiences, and highlight the complexities involved in handling this dataset.

A GASB officer has highlighted significant challenges and inconsistencies in the data provided by different government agencies on USAspending.gov. The officer notes that terms such as "appropriated," "encumbered," and "spent" carry different meanings depending on the agency. For example, HUD operates under annual appropriations, issuing grants yearly, whereas DOT utilizes a trust appropriation lasting five years, which is multiannual. Such discrepancies in reporting can result in misinterpretations if not standardized within the data structure. When agencies like HHS, which operates on an entitlement basis, are added to the mix, the interpretations of these terms vary even more dramatically.

Additionally, the officer shared a personal experience of frustration with the database. A few years ago, their team downloaded the entire database for analysis, aiming to gain insights into grants provided to state and local governments. They found that the vast amounts of data from HHS and Treasury (debt) overshadowed other data, complicating their analyses. Despite efforts to refine the data by removing entries related to HHS and debt and focusing on state and type of government, the challenges persisted. Consequently, the team decided not to use this data to support GASB's recommendations, citing the difficulties in extracting reliable and clear insights from the database. This situation underscores the need for improvements in data management and clarity to support effective and informed decision-making at GASB.

Her parting words remind us that while improvements in data quality are hoped for, the inherent challenges of such a complex dataset should always be considered when designing and applying analytic models. This insight is crucial for our project, as it informs the strategies we must employ to manage and interpret the data effectively, acknowledging the potential traps that lie in disparate data practices across government agencies.

For this project, the decision to focus solely on DHHS data is due to its comprehensive coverage of detailed financial transactions, which minimizes the risk of discrepancies and misinterpretations that could arise from integrating data across different departments. Initially, the DHHS dataset may exhibit heterogeneity; however, once integrated into our outlier detection framework, the data is processed to achieve homogeneity, ensuring that all data elements are standardized and straightforward to interpret. This approach not only simplifies the analysis but also enhances the accuracy and clarity of the insights derived, effectively addressing potential challenges related to data quality.

## VII. Discussion

The reward page on the DHHS website serves as a comprehensive hub where detailed descriptions of various awards are provided. It meticulously catalogs each award by offering in-depth insights into the criteria, eligibility, and significance of the awards. This dedicated section not only enhances transparency by outlining what each award entails but also helps in recognizing and celebrating excellence within the organization or community. Clearly stating what achievements or behaviors are rewarded motivates individuals and teams to strive towards these standards. Moreover, the detailed descriptions ensure that all participants understand the benchmarks for excellence and the values that the organization wishes to promote. This page is instrumental in fostering a culture of acknowledgment and appreciation, where exceptional efforts are publicly recognized, thus boosting morale and encouraging a competitive yet collaborative environment. Whether for academic achievements, professional milestones, or community service, the reward page stands as a testament to the organization's commitment to honoring outstanding contributions in various fields.

In future developments of this project, we plan to integrate Natural Language Processing (NLP) to analyze textual data from the Description section of awards. This section includes key updates and objectives related to the award, providing rich insights into project progression and compliance. Convert key phrases and words from award descriptions into structured, numerical data using NLP techniques like keyword extraction and sentiment analysis. Merge this quantified text data with the existing numerical data, enhancing the outlier detection framework to evaluate both financial and textual information. Apply outlier detection algorithms to this combined dataset to identify anomalies based on discrepancies in textual content relative to typical project updates.

By including textual analysis, the model gains a more comprehensive view of each award, potentially uncovering inconsistencies not visible through financial data alone.

**Conclusion**

In the exploration of unsupervised outlier detection within the domain of audit analytics, specifically focusing on the U.S. Department of Health and Human Services (DHHS) spending data, this study has thoroughly investigated and compared various algorithms to enhance anomaly detection. This study underscores the importance of leveraging advanced analytics to improve audit efficiency and accuracy, especially in handling large-scale, complex datasets. The integration of multiple outlier detection techniques—such as Histogram-based Outlier Score (HBOS), Robust Principal Component Analysis (PCA), Minimum Covariance Determinant (MCD), and K-Nearest Neighbors (KNN)—has proven to be crucial in identifying discrepancies that traditional methods might miss.

The findings from this research emphasize that no single method excels in all scenarios; instead, a hybrid approach that employs various algorithms based on their strengths in specific contexts yields the most robust and reliable results. This strategy not only enhances the detection capabilities of audit systems but also contributes to the design of more effective audit procedures and the development of automated tools for financial oversight.

Future directions for this research include extending the analysis to additional datasets from other government agencies or the private sector to validate and generalize the effectiveness of these models. Further refinement of these techniques, incorporating domain-specific knowledge, could improve their precision. Additionally, exploring the integration of supervised and unsupervised learning methods may offer new insights into enhancing anomaly detection capabilities.

In conclusion, this study has demonstrated the potential of unsupervised outlier detection methods to significantly advance the field of audit analytics. By improving anomaly detection, these techniques can aid in more effective financial oversight and risk management, ultimately contributing to the integrity and transparency of financial systems in an era characterized by voluminous and complex data sets.

**Appendix**

*Figure 7. Outlier Detection Audit Application Framework*

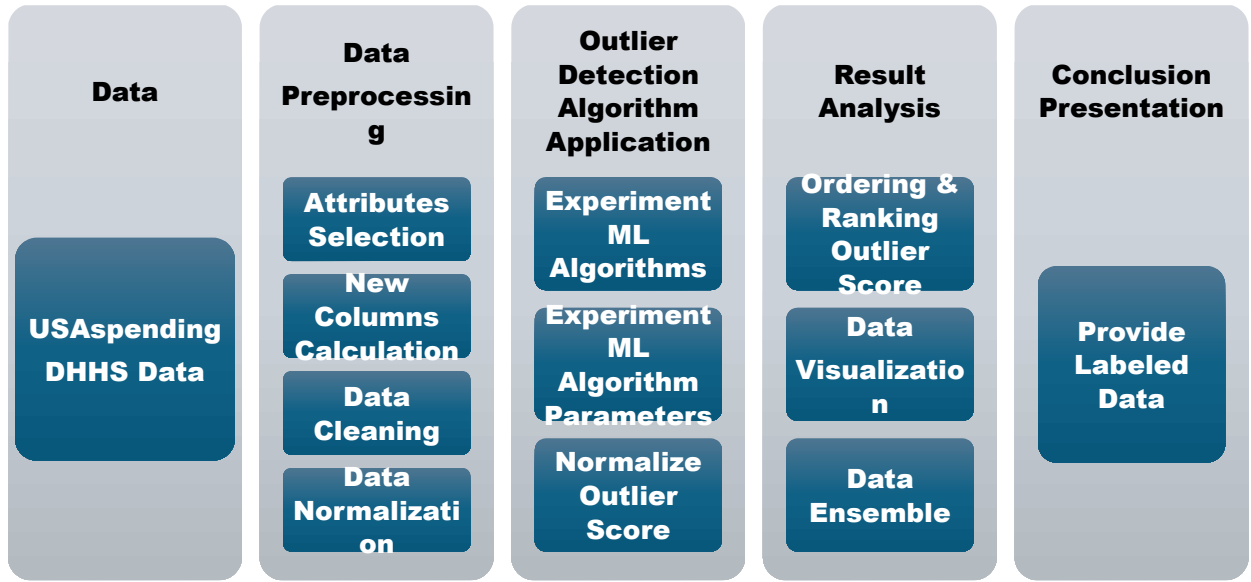